\documentclass[letterpaper, 10 pt, conference]{ieeeconf}
\IEEEoverridecommandlockouts    
\usepackage{graphics}           
\usepackage{times}              
\usepackage{amsmath}            
\usepackage{amssymb}            
\usepackage{graphicx}
\usepackage{algorithm}
\usepackage[noend]{algpseudocode}
\usepackage{booktabs}
\usepackage{color}
\definecolor{instructioncolor}{rgb}{.5,.5,.5}

\usepackage[font=small]{caption}

\def\secref#1{Sec.~\ref{#1}}
\def\figref#1{Fig.~\ref{#1}}
\def\tabref#1{Tab.~\ref{#1}}
\def\eqref#1{Eq.~(\ref{#1})}

\makeatletter
\usepackage{xspace}
\DeclareRobustCommand\onedot{\futurelet\@let@token\@onedot}
\def\@onedot{\ifx\@let@token.\else.\null\fi\xspace}
 
\def\ie{i.e\onedot}

\def\etal{{et al}\onedot}
\makeatother

\def\etalcite#1{\etal~\cite{#1}}

\usepackage{array}
\newcolumntype{L}[1]{>{\raggedright\let\newline\\\arraybackslash\hspace{0pt}}m{#1}}
\newcolumntype{C}[1]{>{\centering\let\newline\\\arraybackslash\hspace{0pt}}m{#1}}
\newcolumntype{R}[1]{>{\raggedleft\let\newline\\\arraybackslash\hspace{0pt}}m{#1}}

\def\argmin{\mathop{\rm argmin}}

\newcommand{\RR}{\mathbb{R}}

\newcommand{\mybold}[1]{\mbox{\boldmath$#1$}}

\newcommand{\vv}[1]{{\mybold #1}} 

\newcommand{\m}[1]{{\mbox{{\sffamily\slshape{#1\/}}}}}

\newcommand{\q}[1]{{{\bf #1}}}

\newcommand{\mq}[1]{{\mbox{{\sffamily{#1}}}}}

\newcommand{\tr}[0]{\sf T}              

\usepackage{flushend}
\usepackage{siunitx}
\usepackage{multirow}
\usepackage{pifont}
\usepackage{amssymb}
\usepackage{amsmath}
\usepackage{subfigure}
\usepackage{tabularx}
\usepackage{comment}

\usepackage{stmaryrd}
\usepackage{multirow}
\usepackage{pifont}

\makeatletter
\let\NAT@parse\undefined
\makeatother
\usepackage{hyperref}  

\newcommand{\rgbd}{\mbox{RGB-D}\ } 
\newcommand{\cmark}{\ding{51}}%
\newcommand{\xmark}{\ding{55}}%

\newcommand{\equsize}{\normalsize} 

\renewcommand{\and}{\hspace{1.0cm}} 

\title{\LARGE \bf  Panoptic Mapping with Fruit Completion \\ and Pose Estimation for Horticultural Robots}
\author{Yue Pan \and Federico Magistri \and Thomas Läbe \and Elias Marks \\ Claus Smitt \and  Chris McCool \and Jens Behley \and Cyrill Stachniss}

\begin{document}

\twocolumn[{%
\renewcommand\twocolumn[1][]{#1}%

\maketitle
\thispagestyle{empty}
\pagestyle{empty}

\begin{center}
    \centering
    \vspace{-2em}
    \captionsetup{type=figure}
    \includegraphics[width=0.95\linewidth]{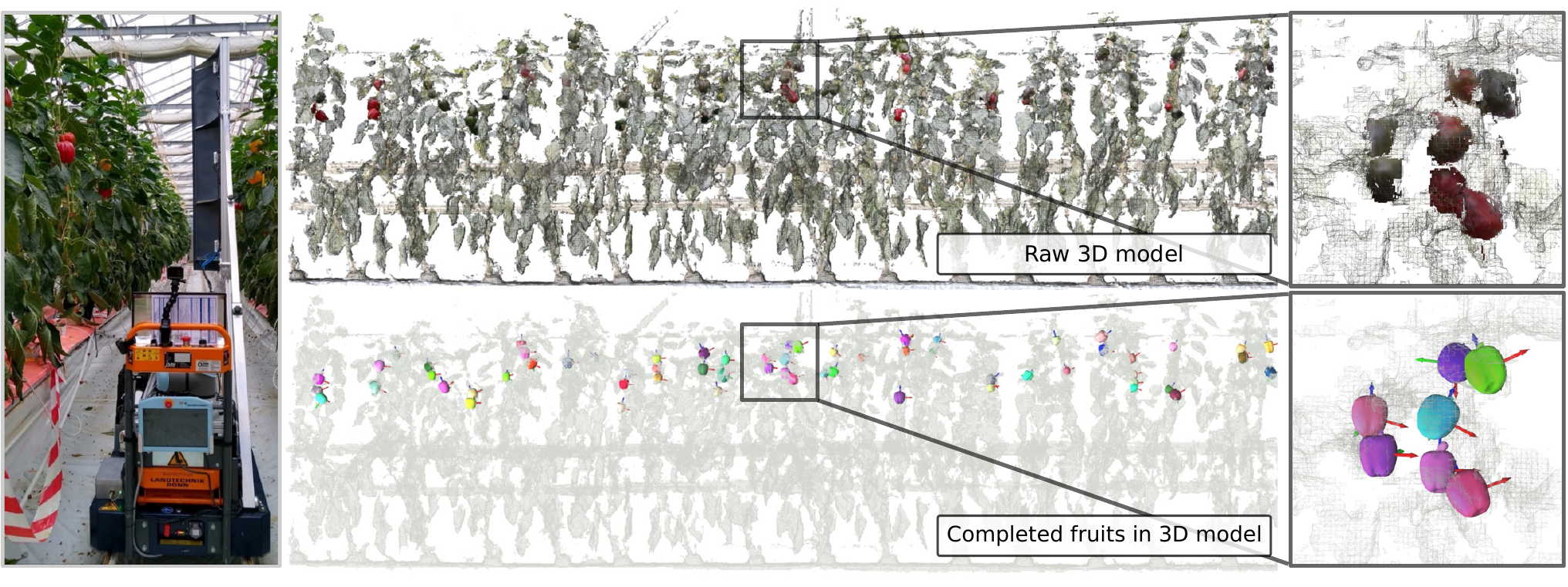}
    \setlength{\abovecaptionskip}{1pt}
    \captionof{figure}{Our method is able to build a multi-resolution panoptic map (top) of a challenging commercial greenhouse environment online using a mobile horticultural robot equipped with RGB-D cameras (left). Furthermore, our method manages to jointly estimate the complete shape and pose of each fruit in the panoptic map (bottom).}
    \label{fig:mot}
    \vspace{1mm}
\end{center}%
}]

\makeatletter{\renewcommand*{\@makefnmark}{}
\footnotetext{All authors are with the University of Bonn, Germany. Cyrill Stachniss is additionally with the Department of Engineering Science at the University of Oxford, UK, and with the Lamarr Institute for Machine Learning and Artificial Intelligence, Germany.}\makeatother
\footnotetext{This work has partially been funded 
  by the Deutsche Forschungsgemeinschaft (DFG, German Research Foundation) under Germany's Excellence Strategy, EXC-2070 -- 390732324 -- PhenoRob and under STA~1051/5-1 within the FOR 5351~--~459376902~(AID4Crops).}\makeatother
}

\setlength{\textfloatsep}{1.3em}
\setlength{\dbltextfloatsep}{1.3em}

\thispagestyle{empty}
\pagestyle{empty}

\begin{abstract}
  Monitoring plants and fruits at high resolution play a key role in the future of agriculture. Accurate 3D information can pave the way to a diverse number of robotic applications in agriculture ranging from autonomous harvesting to precise yield estimation. Obtaining such 3D information is non-trivial as agricultural environments are often repetitive and cluttered, and one has to account for the partial observability of fruit and plants.
  In this paper, we address the problem of jointly estimating complete 3D shapes of fruit and their pose in a 3D multi-resolution map built by a mobile robot. 
  To this end, we propose an online multi-resolution panoptic mapping system where regions of interest are represented with a higher resolution. We exploit data to learn a general fruit shape representation that we use at inference time together with an occlusion-aware differentiable rendering pipeline to complete partial fruit observations and estimate the 7 DoF pose of each fruit in the map.
  The experiments presented in this paper, evaluated both in the controlled environment and in a commercial greenhouse, show that our novel algorithm yields higher completion and pose estimation accuracy than existing methods, with an improvement of 41\% in completion accuracy and 52\% in pose estimation accuracy while keeping a low inference time of 0.6\,s in average.
\end{abstract}

\section{Introduction}
\label{sec:intro}

To feed an ever-growing world population, the whole agricultural sector needs to increase its productivity while reducing its negative impact on the environment. Autonomous robots have the potential to support addressing both issues.
For example, robots can automate labor-intensive tasks such as harvesting~\cite{arad2020jfr, lehnert2017ral}, weeding~\cite{mccool2018ral}, or pruning~\cite{botterill2017jfr}. 
They can also provide plant-specific treatment of herbicides and pesticides~\cite{ahmadi2022iros} reducing the required amount of agro-chemicals. 
Thus increasing the likelihood of meeting the population demands in producing food, feed, fiber, and fuel and at the same time decreasing the use of agro-chemicals.

Robots can continuously monitor orchards or arable fields to detect early stages of plant stress~\cite{yi2020sensors}, support phenotyping activities~\cite{zermas2020cea}, and provide detailed yield estimates~\cite{kierdorf2022fai}.  
Robots working in arable fields or horticulture environments can seldomly observe the whole scene due to the cluttered nature of the environments. This means that data obtained with any agricultural robot is partial and incomplete.
In this paper, we address the problem of building multi-resolution 3D maps of such scenes and estimating the non-visible parts of fruit to obtain more complete 3D models. \figref{fig:mot} depicts an example of the resulting map together with the 3D shape and pose estimated for each fruit.

\begin{figure*}[ht]
  \centering  
  \includegraphics[width=0.96\linewidth]{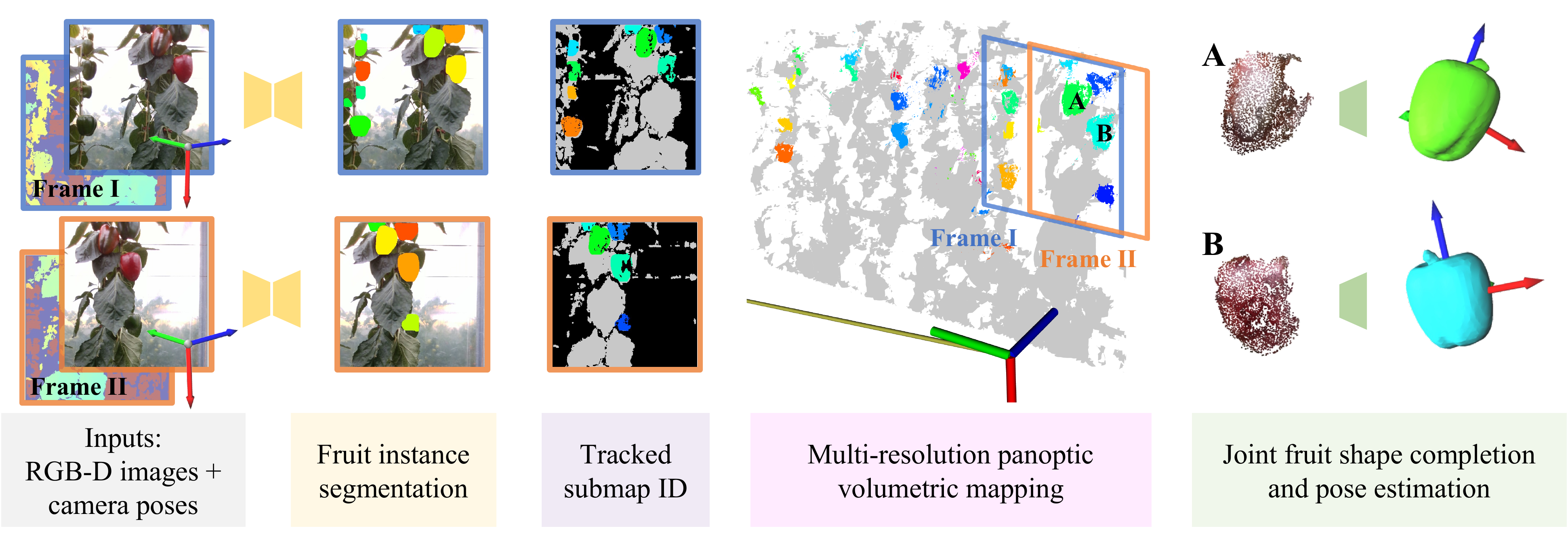}
  \caption{Overview pipeline of the proposed mapping system: the input to the mapping system is a stream of the \rgbd images collected by the horticultural mobile robot. We predict the fruit instance using Mask R-CNN~\cite{he2017iccv-mr}. Then we conduct panoptic volumetric mapping by tracking the instance lables as the temporal consistent submap IDs. For each fruit submap, we jointly estimate the complete shape and pose from the submap points and the corresponding images using offline learned shape prior and the differentiable rendering pipeline.}
  \label{fig:overview}
  \vspace{-10pt}
\end{figure*}
To recover the shapes of occluded objects, a typical solution is the usage of templates whose deformations are able to represent different instances of the same category. A template can be represented as a 3D triangular mesh~\cite{magistri2021icra} or encoded in the weights of a neural network~\cite{park2019cvpr}. In recent years, templates have been used to recover shapes of human bodies~\cite{gall2009cvpr} and hands~\cite{tagliasacchi2015cgf}, but also in the agricultural context to estimate shapes of fruits~\cite{magistri2022ral-iros} and plants~\cite{marks2022icra}.

The main contribution of this paper is a novel method to jointly estimate the 3D shape of fruits and their pose. We build multi-resolution maps in which we place the predicted 3D shapes of fruits correctly posed in a globally consistent representation. Additionally, we exploit high-resolution 3D data to encode a general fruit representation into the weights of a neural network. In this way, we can recover details of the fruit's shape with a low inference time during operations.

In sum, we make three key claims:
our approach is able to jointly
(i) predict the 3D shape of fruits even under substantial occlusions in real commercial greenhouse environments and
(ii) estimate the pose of each fruit in the 3D map.
(iii) Additionally, our multi-resolution map representation yields substantial improvement in the shape completion task over fixed-resolution maps.
These claims are backed up by our experimental evaluation.

Our open-source implementation is publicly available at: \url{https://github.com/PRBonn/HortiMapping}.






\section{Related Work}
\label{sec:related}


In recent years, agricultural robotics has become an increasingly popular research area due to the aforementioned challenges and the prospect of deploying such robots for more efficient and sustainable agri- and horticulture. More specifically, in the horticulture context we have seen deployed robotic systems for monitoring~\cite{smitt2021icra} and harvesting~\cite{arad2020jfr,lehnert2017ral}. In both cases, a fundamental build block of such robots is an instance segmentation network~\cite{he2017iccv-mr} that can robustly segment fruits~\cite{halstead2020dicta,mccool2016icra}, peduncles~\cite{sa2017icra}, and plants~\cite{halstead2021fps}, often starting from 2D images.
Neural networks can also be used to estimate ripeness~\cite{halstead2018ral} and poses~\cite{wagner2021icra} of fruits.
Based on such networks, it is possible to build semantically-aware 3D maps~\cite{rosinol2020icra}. However, in the agricultural context, classical volumetric mapping pipelines, such as KinectFusion~\cite{newcombe2011ismar} or Voxblox~\cite{oleynikova2017iros}, yield incomplete fruits representations as they do not deal explicitly with the presented occlusions. Instead, we build upon a multi-resolution map from prior work~\cite{pan2022iros,schmid2022icra} and propose a pipeline to jointly estimate the pose and the complete shape of fruits in the 3D map resulting in a complete and better representation of fruits.

Recently, a variety of works tackle the problem of estimating the shape of non-visible fruits parts of plants or fruits. Such works can be generally divided into three categories: geometry-based, mesh-based, and deep learning-based approaches.
In the first category, a closed-form geometric model is fit into the collected data. Marangoz~\etalcite{marangoz2022case} use such an approach for fruit monitoring, while Lehnert~\etalcite{lehnert2016icra} show that fitting a superellipsoid improves robotic grasping performances. These approaches estimate complete shapes quickly but the model used cannot represent details in the 3D shapes. 

In the second category, the model is represented by a 3D mesh that represents the general appearance of a target object. We used this approach to estimate the shapes of plants~\cite{magistri2021icra} and leaves~\cite{marks2022icra}. While such methods can provide precise reconstructions, a shortcoming is the high inference time required for obtaining the final mesh. This is often not practical for real-world operations. 

In the last category, the general appearance of an object is encoded in the weights of a neural network that produces either point clouds~\cite{yuan2018threedv} or triangular meshes~\cite{park2019cvpr}. In our previous work~\cite{magistri2022ral-iros}, we use DeepSDF~\cite{park2019cvpr} to learn a prior over fruits shapes and map a single \rgbd frame to the network's latent space to avoid the online optimization. While such approaches provide plausible 3D shapes at a low inference time, they lack the ability to estimate 7 degrees of freedom (DoF) poses and the performance declines when processing the fruits with non-canonical poses in the real world. The effectiveness of DeepSDF-based methods~\cite{magistri2022ral-iros} on in-the-wild settings is thus limited by the requirement of having the input partial shapes in the same canonical pose and scale as in the training set. 

Inspired by the recent works in the computer vision community including Frodo~\cite{runz2020cvpr} and DSP-SLAM~\cite{wang2021threedv} we propose a novel method to jointly estimate the shape and the pose of fruits with a low inference time. We combine a completion network based on a neural shape prior and an occlusion-aware differentiable rendering pipeline. In this way, the shape completion and reconstruction are still conducted in the canonical pose of the fruit with high quality while obtaining the transformation to the world coordinate system and allowing for its embedding in a 3D model of the greenhouse and its plants.




\section{Our Approach to Fruit Mapping}
\label{sec:main}


In this paper, we study the problem of panoptic volumetric mapping for horticultural applications while estimating complete 3D shapes including scale and 7 DoF pose of fruits from a sequence of \rgbd frames.

As shown in \figref{fig:overview}, we build a panoptic volumetric map that decomposes the scene into the background and individual foreground submaps. The background submap is represented at a lower resolution than the foreground submaps to ensure accurate and high-fidelity reconstructions of the fruits. Based on the foreground submaps integrating partial observations, we estimate complete shapes and 7 DoF poses of each individual fruit using a deep neural network that jointly estimates the shape of the fruit in a canonical coordinate system and a transformation into the world coordinate frame.

\subsection{Multi-resolution Panoptic Mapping}
\label{sec:mapping}
The input to our mapping system is a stream of the \rgbd images collected by a horticultural mobile robot. The pose of the robot can be estimated by wheel odometry or a tracking camera and refined online with a standard \rgbd odometry system. 
We train a Mask R-CNN~\cite{he2017iccv-mr} model to perform online instance segmentation of the fruits based on the RGB information resulting in per-fruit instance masks $\mathcal{M}$. In line with Panoptic Multi-TSDFs~\cite{schmid2022icra}, we assign a submap $\mathcal{S}$ to each panoptic entity, either the thing (fruit) instance or the stuff (background). For each submap, we apply a non-projective TSDF integration~\cite{pan2022iros} as well as a mesh reconstruction based on marching cubes algorithm~\cite{lorensen1987siggraph} incrementally using the \rgbd image stream and the estimated robot's poses. 

Meanwhile, we track the segmented instances as temporal consistent submap IDs through a data association. We project the mesh vertices of active submaps within the visual frustum onto the image planes to generate submap masks for newly collected image frames. Based on the intersection over union~(IoU) ratio between the rendered and predicted masks and the difference between the rendered and measured depth, we either assign a segmented instance to an existing submap or allocate a new submap for it. 

\begin{figure}
  \centering  
  \includegraphics[width=0.95\linewidth]{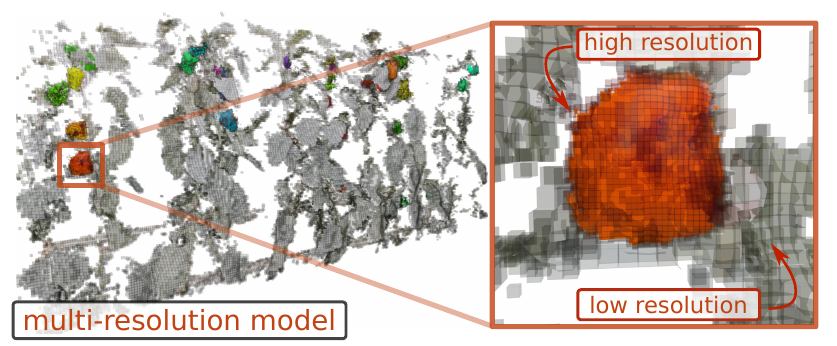}
  \caption{On the left, we show an example of our multi-resolution volumetric map consisting of a lower resolution background (with larger voxel size) and several higher resolution fruit submaps (with smaller voxel size). On the right, we zoom in to show the high resolution submap representing a fruit.}
  \label{fig:multires}
  \vspace{2mm}
  \centering  
  \includegraphics[width=0.97\linewidth]{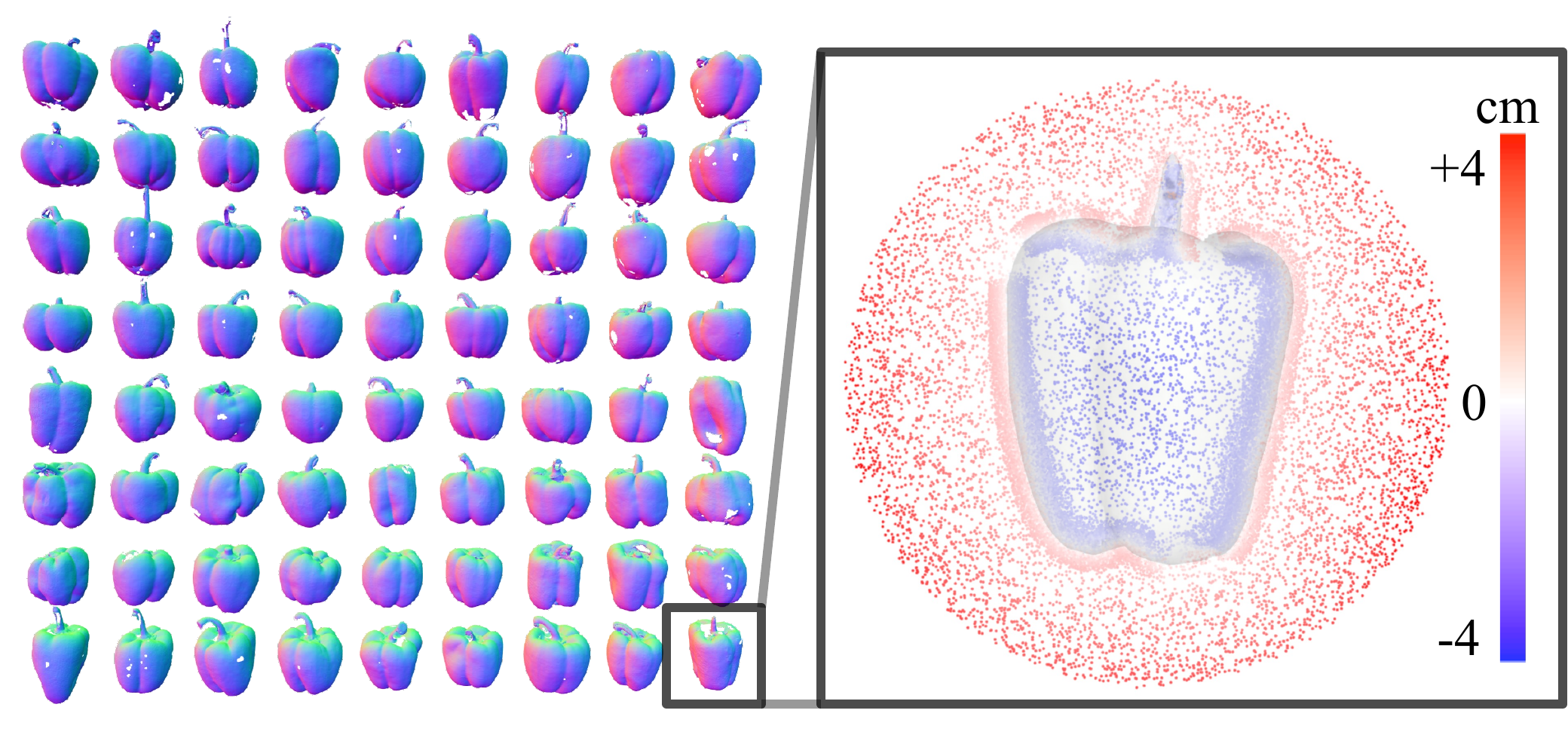}
  \caption{On the left, we show examples of the sweet pepper models with a canonical pose measured using a sub-millimeter precision laser scanner. On the right, for one example sweet pepper, we show both the close-to-surface sample points and the uniform free space sample points with their color representing the SDF value label. We use these sample points and their SDF value labels to train the DeepSDF model and learn the shape prior of the specific fruit species (in this case the sweet pepper). }
  \label{fig:deep_sdf_training}
\end{figure}

As shown in \figref{fig:multires}, our volumetric mapping allows the fruits 
 to be reconstructed
at a small voxel size ($3\,\mathrm{mm}$) and with high accuracy, while using
larger voxel sizes ($1\,\mathrm{cm}$) to efficiently map the background or less
relevant classes such as leaves and stems. 
Therefore, the final panoptic map consists of a low-resolution submap for the background and high-resolution submaps for each fruit instance. 

Once a fruit submap is not observed for $G$ frames consecutively, it is not updated anymore, marked as frozen, and used for joint shape completion and pose estimation.

\subsection{Fruit Completion and Pose Estimation in the Wild}
\label{sec:optimization}

By exploiting the fruit shape prior of a pre-trained DeepSDF~\cite{park2019cvpr} model and an occlusion-aware differentiable rendering technique, for each frozen fruit submap $\mathcal{S}$ from the panoptic volumetric map in \secref{sec:mapping}, we aim to jointly estimate its latent shape code $\vv{z}$ and 7 DoF transformation $\mq{T}_{o w} \in \operatorname{Sim}(3)$ from the world coordinate system to the fruit's canonical coordinate system. $\mq{T}_{o w}$ is represented by an homogeneous transformation matrix with $s \in \RR$, $\m{R} \in\RR^{3\times3}$, $\vv{t} \in \RR^3$ representing the scale, rotation,  and translation, respectively.
We sample a point cloud $\mathcal{P}_\mathcal{S}$ from the mesh of each instance submap $\mathcal{S}$. Additionally, from all the images in which $\mathcal{S}$ is visible, we get the 2D mask $\mathcal{M}$ predicted by Mask R-CNN, the extended 2D bounding box $\mathcal{B}$ with a padding of $h$ pixels on each side.

\textbf{DeepSDF pretraining:} The DeepSDF~\cite{park2019cvpr} model takes as input a query position $\vv{x} \in \RR^3$ and a latent shape code $\vv{z} \in \RR^{C}$, and predicts the SDF value $v \in \RR$ at $\vv{x}$ through a decoder network $D_{\theta}$ as: 
  $v = D_\theta(\vv{x}, \vv{z})$.
With a pre-trained DeepSDF decoder $D_{\theta}$ and optimized shape code $\vv{z}$, we compute a dense SDF volume by querying it at a regular 3D grid of points, which we use for a complete mesh reconstruction via marching cubes~\cite{lorensen1987siggraph}. 

%
In line with Magistri \etal~\cite{magistri2022ral-iros}, we aim at having an accurate and complete model of the fruits predicted by $D_\theta$ for which we use high-resolution 3D laser scans collected in a controlled laboratory environment using a sub-millimeter accurate Perceptron V5 laser scanner and a Romer Infinite measuring arm.
We train a DeepSDF model for each type of fruit using the measured 3D scans and instead of only sampling points close to the surface along the normal direction~\cite{park2019cvpr,magistri2022ral-iros}, we also sample points uniformly in a sphere surrounding the object as shown in \figref{fig:deep_sdf_training}. By doing so, the model better represents the free space, which is beneficial for differentiable rendering and pose estimation. Regarding model training, we adopt a latent code size of $C=32$ and adhere to the network architecture and hyperparameter configurations specified in the original work of DeepSDF~\cite{park2019cvpr}.

\begin{figure}
  \centering  
  \includegraphics[width=0.84\linewidth]{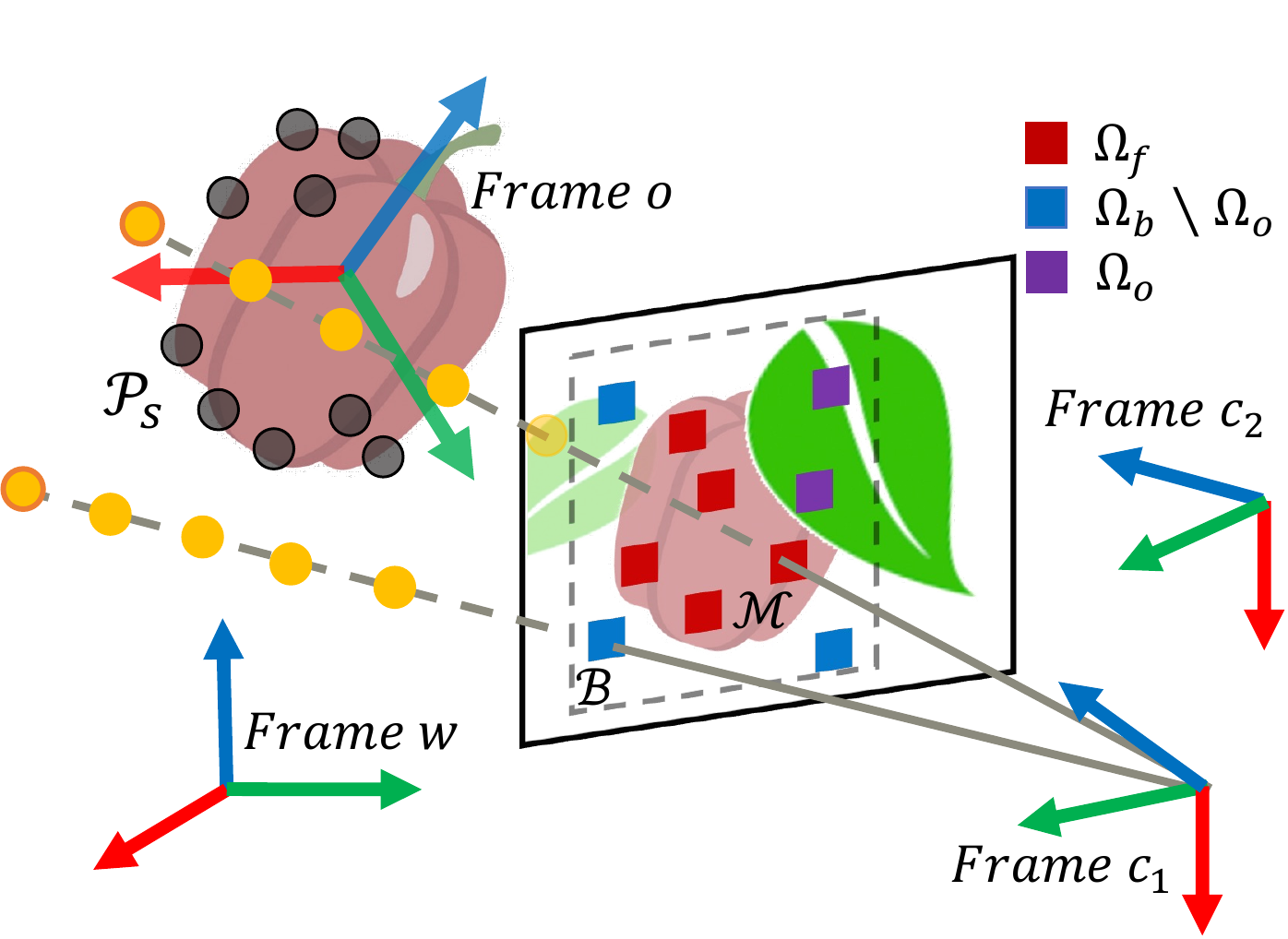}
  \caption{Geometric relationships for the joint optimization of the shape $\vv{z}$ and the pose $\mq{T}_{o w}$ of a target fruit using the submap point cloud in the world frame $w$ and the image collected in the camera frame $c$. The black points represent the sampled points from the point cloud $\mathcal{P}_\mathcal{S}$ of the target fruit submap. The orange points represent the sample points on each ray corresponding to each sampled pixel. The red, blue, and purple squares represent the sampled pixels from the masked foreground region $\mathcal{M}$, the unmasked background region in the extended bounding box $\mathcal{B}$ and the potential occluded background region, respectively.}
  \label{fig:losses}
  \vspace{-6pt}
\end{figure}

\textbf{Surface consistency loss:} Our first objective is to achieve precise alignment between the reconstructed fruit and the fused 3D observations from the \rgbd sensor. To accomplish this, as shown in~\figref{fig:losses}, we aim to keep the points from the target fruit submap's point cloud $\mathcal{P}_\mathcal{S}$ close to the iso-surface of the SDF predicted by $D_{\theta}$, which minimizes the following surface reconstruction loss $\mathcal{L}_{\mathrm{s}}$ given by:
\begin{align}
  \mathcal{L}_{\mathrm{s}} &= \frac{1}{\left|\mathcal{P}_\mathcal{S}\right|} \sum_{\q{p}^{w} \in \mathcal{P}_\mathcal{S}} D_\theta^2\left(\mq{T}_{o w}\q{p}^{w}, \vv{z}\right).
\end{align}

\textbf{Occlusion-aware differentiable rendering:} 
As we use depth images and instance segments, we additionally propose a depth rendering loss and a mask rendering loss using differentiable SDF rendering.
As shown in~\figref{fig:losses}, for each image corresponding to a visible fruit submap, we sample $G_f$ foreground pixels $\boldsymbol{\Omega}_f$ from the mask $\mathcal{M}$ and $G_b$  background pixels $\boldsymbol{\Omega}_b$ from the unmasked region $\mathcal{B}\setminus\mathcal{M}$ in the extended bounding box $\mathcal{B}$. For each sampled pixel \mbox{$\vv{u} \in \boldsymbol{\Omega} = \boldsymbol{\Omega}_f \cup \boldsymbol{\Omega}_b$}, we calculate the corresponding ray $\q{r}$ in the camera coordinate system. We then sample $N+1$ points with a fixed interval $\Delta d$ on each ray, resulting in per sample depths $d_i$:
\begin{align}
d_i &= d_{\min}+(i-1) \Delta d, \qquad i=1,\ldots, N+1,
\end{align}
along the ray $\q{r}$ from the projection center to the fruit, where $d_{\min}$ and $\Delta d$ are determined by the distance from the camera to the submap's bounding box center and the approximate fruit size. By transforming each sample point to the world coordinate system, we obtain:
\begin{align}
\q{p}^{w}_{i} = \mq{T}_{w c}\left(d_i \mq{K}^{-1} \q{u}\right),
\end{align}
where $\mq{K}$ and $\mq{T}_{w c}$ are the camera intrinsic and extrinsic matrices, respectively. Then,  the SDF prediction $v_i$ at the sample point $\q{p}^{w}_{i}$ is given by:
\begin{align}
  v_{i} &= D_\theta(\mq{T}_{o w}\q{p}^{w}_{i}, \vv{z}),
\end{align}
which can then be converted to an estimated occupancy probability $o_{i}$ by a logistic function with a surface noise threshold $\sigma$:
\vspace{-1em}
\begin{align}
  o_{i}=\frac{1}{1+e^{v_{i} / \sigma}}.
\end{align}

Then, the so-called ray termination weight $\alpha_i$ for each sample point can be calculated from the occupancy probability as:
\vspace{-1em}
{\equsize 
\begin{equation}
\alpha_i = \begin{cases}o_i \prod_{j=1}^{i-1}\left(1-o_j\right) & i=1, \ldots, N \\ \prod_{j=1}^{N}\left(1-o_j\right) & i=N+1. \end{cases}
\end{equation}
}

Note that $\sum_{i=1}^{N+1} \alpha_i = 1$ always holds. We can then get the rendered mask $\hat{M}$ and depth $\hat{D}$ by integrating over all the samples along the ray as: 
{\equsize 
\begin{equation}
 \hat{M}=\sum_{i=1}^N \alpha_i, \quad \hat{D}=\sum_{i=1}^{N+1} \alpha_i d_i. 
\end{equation}
}

The depth rendering loss $\mathcal{L}_{\mathrm{d}}$ and mask rendering loss $\mathcal{L}_{\mathrm{m}}$ for $K$ image frames observing the fruit submap are calculated by comparing the rendered result $\hat{D}$ and $\hat{M}$ with the depth camera's measurement $D$ and the binary mask $M$ of the target fruit for each of the $G$ sampled pixels as:
\begin{eqnarray}
  \mathcal{L}_{\mathrm{d}}=\frac{1}{KG} \sum_{k}\sum_{\mathbf{u} \in (\boldsymbol{\Omega}^k \setminus \boldsymbol{\Omega}^k_o)}\left(\hat{D}_{\mathbf{u}}-D_{\mathbf{u}}\right)^2,\\
  \mathcal{L}_{\mathrm{m}}=\frac{1}{KG} \sum_{k}\sum_{\mathbf{u} \in (\boldsymbol{\Omega}^k \setminus \boldsymbol{\Omega}^k_o)}\left(\hat{M}_{\mathbf{u}}-M_{\mathbf{u}}\right)^2.
  \label{equ:mask}
\end{eqnarray}

For background pixels, if $\hat{D}_{\mathbf{u}}-D_{\mathbf{u}}>d_o$, where $d_o$ is a small threshold, the pixel is regarded as lying in a potential occluded region of the fruit caused by leaves or other fruits, as shown in~\figref{fig:losses}. Such occluded pixels $\boldsymbol{\Omega}_o \subset \boldsymbol{\Omega}_b$ are not taken into account in the rendering loss $\mathcal{L}_{\mathrm{d}}$ and $\mathcal{L}_{\mathrm{m}}$. For the rest of background pixels $\boldsymbol{\Omega}_b \setminus \boldsymbol{\Omega}_o$, we take the termination depth $d_{\max}=d_{\min}+N\Delta d$ as the virtual depth measurements for depth rendering loss calculation so that we can enforce the silhouette consistency at the target fruit.

\subsection{Optimization}
With an additional shape code regularization term \mbox{$\mathcal{L}_{\mathrm{r}} = \|\vv{z}\|^2$}, we use as our final loss function:
\begin{equation}
\mathcal{L}=w_{\mathrm{s}} \mathcal{L}_{\mathrm{s}}+w_{\mathrm{d}} \mathcal{L}_{\mathrm{d}}+w_{\mathrm{m}} \mathcal{L}_{\mathrm{m}}+w_{\mathrm{r}}\mathcal{L}_{\mathrm{r}},
\label{equ:all_loss}
\end{equation}
\noindent where $w_{\mathrm{s}}$, $w_{\mathrm{d}}$, $w_{\mathrm{m}}$, $w_{\mathrm{r}}$ are the weights for each loss term. 

Our goal is to solve $\vv{\xi}_{o w}^*,\vv{z}^*=\argmin \mathcal{L}$, where  $\vv{\xi}_{o w} \in \mathbb{R}^7$ is the corresponding $\mathfrak{s i m}(3)$ Lie Algebra of $\mq{T}_{o w}$. Instead of using first-order optimization such as gradient descent, we use  Levenberg-Marquardt with analytical Jacobians for faster and more stable convergence. The latent shape code $\vv{z}$ is initialized as $\vv{0}_{C}$ while $\vv{\xi}_{o w}$ is initialized as an identity rotation, scaling of 1, and a translation from the bounding box center of the submap point cloud to the origin. For each iteration, with a damping parameter $\lambda$, the increment $\delta \vv{x}$ to the estimated parameter vector $\left[\vv{\xi}_{o w}, \vv{z} \right]^{\tr}$ is given by:
\begin{equation} 
  \delta \vv{x} = \left[\delta \vv{\xi}_{o w}, \delta \vv{z} \right]^{\tr} = (\m{H} + \lambda \mathrm{diag}(\m{H}))^{-1} \vv{g}
\label{equ:lm}
\end{equation}

The approximate Hessian matrix is given by $\m{H}=\m{J}^{\tr} \m{P} \m{J}$ and the gradient of the target function is $\vv{g}=\m{J}^{\tr} \m{P} \vv{b}$, where $\m{J}$, $\m{P}$, $\vv{b}$ are the Jacobian matrix, weight matrix and residual vector, respectively.

Since both, the submap point cloud and the camera depth measurements are noisy, we apply a Huber robust kernel for the surface reconstruction and depth rendering residual resulting in the weight $w^{h}_{i}$ for each observation:
\begin{equation}
  w^{h}_{i}= \begin{cases}1 &, \text{if } \left|e_i\right|\leq\tau \\ \tau /\left|e_i\right| & \text{, otherwise} \end{cases},
\end{equation}
where $e_i$ and $\tau$ are the corresponding residual and kernel threshold, respectively. The weight matrix is formulated as $\m{P}=\mathrm{diag}\left(w_{s}w_{1}^{h}, \cdots, w_{d}w_{n}^{h}\right)$ with both the loss specific weight $w_{\mathrm{s}}$, $w_{\mathrm{d}}$ and the Huber weight $w^{h}$.

The residual and Jacobian of the surface reconstruction term for each submap point is given by:
{\equsize 
\begin{eqnarray}
\vv{b}_s=-D_\theta\left(\vv{p}^{o}, \vv{z}\right), \qquad 
\m{J}_s=\frac{\partial D_\theta\left(\vv{p}^{o}, \vv{z}\right)}{\partial\left[\vv{\xi}_{o w}, \vv{z}\right]^{\tr}}.
\end{eqnarray}
}
\vspace{-0.5em}

Applying the chain rule, we obtain the derivative of $\vv{\xi}_{o w}$:
{\equsize
\begin{eqnarray}
\frac{\partial D_\theta\left(\vv{p}^{o}, \vv{z}\right)}{\partial \vv{\xi}_{o w}}=\frac{\partial D_\theta\left(\vv{p}^{o}, \vv{z}\right)}{\partial \vv{p}^{o}} \frac{\partial \vv{p}^{o}}{\partial \vv{\xi}_{o w}},\\
\frac{\partial \vv{p}^{o}}{\partial \vv{\xi}_{o w}} = \frac{\partial (\mq{T}_{o w}\q{p}^{w})}{\partial \vv{\xi}_{o w}} = \left[\begin{array}{lll}
    \m{I}_{3\times3} & -\vv{p}_{\times}^{o} & \vv{p}^{o}
    \end{array}\right],
\end{eqnarray}
}
\vspace{-0.5em}

\noindent where $_{\times}$ refers to the skew symmetric matrix of a vector. Note that both, $\frac{\partial D_\theta\left(\vv{p}^{o}, \vv{z}\right)}{\partial\vv{p}^{o}}$ and $\frac{\partial D_\theta\left(\vv{p}^{o}, \vv{z}\right)}{\partial\vv{z}}$ can be obtained through automatic differentiation of the DeepSDF model $D_\theta$.
The residuals of the rendering term for each sampled pixel are simply:
\vspace{-0.5em}
\begin{equation}
\vv{b}_{d} = \hat{D} - D, \qquad \vv{b}_{m} = \hat{M} - M,
\end{equation}
\noindent and the Jacobians can also be obtained using the chain rule:
{\equsize
\begin{flalign}
  \m{J}_d & =\sum_{i=1}^N \frac{\partial \hat{D}}{\partial o_i} \frac{\partial o_i}{\partial v_i} \frac{\partial D_\theta\left(\vv{p}^{o}_{i}, \vv{z}\right)}{\partial\left[\vv{\xi}_{o w}, \vv{z}\right]^{\tr}},\\
  \m{J}_m & =\sum_{i=1}^N \frac{\partial \hat{M}}{\partial o_i} \frac{\partial o_i}{\partial v_i} \frac{\partial D_\theta\left(\vv{p}^{o}_{i}, \vv{z}\right)}{\partial\left[\vv{\xi}_{o w}, \vv{z}\right]^{\tr}},
\end{flalign}
}

\noindent in which $\frac{\partial D_\theta\left(\vv{p}^{o}_{i}, \vv{z}\right)}{\partial\left[\vv{\xi}_{o w}, \vv{z}\right]^{\tr}}$ is in the same form as $\m{J}_{s}$ and the other three derivative terms are derived as:
{\equsize
\begin{flalign}
\frac{\partial \hat{D}}{\partial o_i} &= \frac{\Delta d}{1-o_i}\sum_{j=i}^N \prod_{k=1}^j (1-o_k), \\
\frac{\partial \hat{M}}{\partial o_i} &= \prod_{j=1, j \neq i}^N\left(1-o_j\right), \\
\frac{\partial o_i}{\partial v_i} &= \frac{o_i(1-o_i)}{\sigma}.
\end{flalign}
\vspace{-0.5em}
}

Lastly, the residual and Jacobian of the shape code regularization term are:
\begin{align}
  \vv{b}_r &=-\vv{z}\\
  \m{J}_r &= \left[\begin{array}{ll}
    \frac{\partial \vv{z}}{\partial \boldsymbol{\xi}_{o w}} & \frac{\partial \vv{z}}{\partial \vv{z}} 
    \end{array}\right]=\left[\begin{array}{ll} \vv{0}_{1\times7} & \vv{1}_{1\times C} \end{array}\right]
\end{align}

With all Jacobians and residuals available, we are able to solve \eqref{equ:lm} and update:
\begin{equation}
\left[\vv{\xi}_{o w}^{(t+1)}, \vv{z}^{(t+1)} \right]^{\tr} = \left[\vv{\xi}_{o w}^{(t)}, \vv{z}^{(t)} \right]^{\tr} + \delta \vv{x}^{(t)},
\end{equation}
until convergence. 
After convergence, the complete fruit model can be reconstructed using marching cubes with the optimized $\vv{z}^{*}$ at 3D grid queries in the fruit's canonical coordinate system. The reconstruction can then be transformed into the world coordinate system using $\mq{T}_{o w}^{*} = \exp (\vv{\xi}_{o w}^{*})$.  

\begin{figure}
  \centering  
  \includegraphics[width=0.98\linewidth]{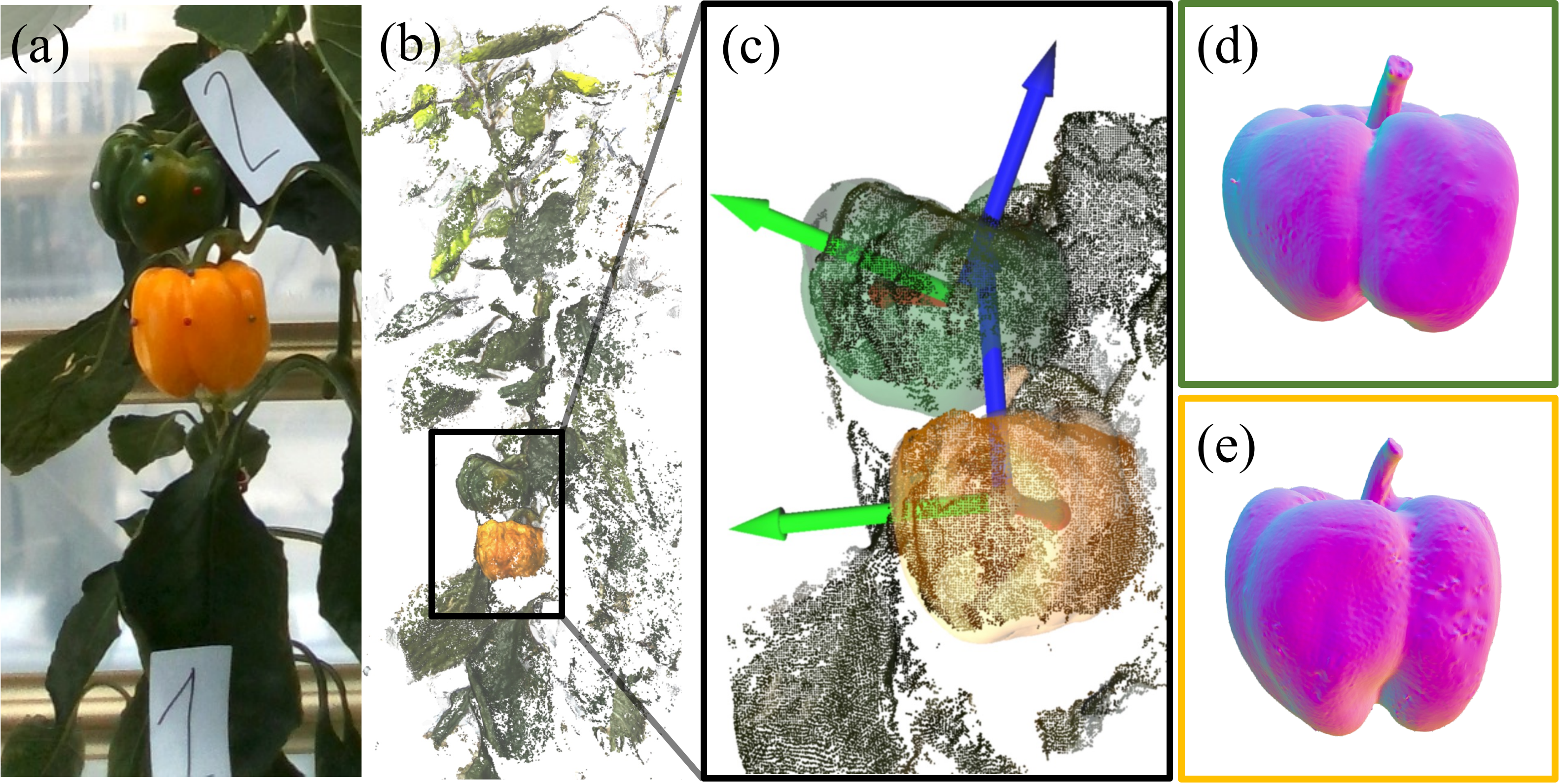}
  \caption{Example from our greenhouse dataset: (a) shows the image of two sweetpeppers with the ID \#1 and \#2. (b) shows the point cloud generated from the camera of the robot by offline bundle adjustment. (c) shows the ground truth complete shape models with the ground truth pose manually aligned with the bundle adjusted point cloud. (d) and (e) show the ground truth shape models of sweetpepper \#2 and \#1 obtained by harvesting the fruits and scanning them in the laboratory with a high-precision laser scanner.}
  \label{fig:dataset}
\end{figure}

\section{Experimental Evaluation}
\label{sec:exp}

\begin{figure*}[ht]
  \centering  
  \includegraphics[width=0.92\linewidth]{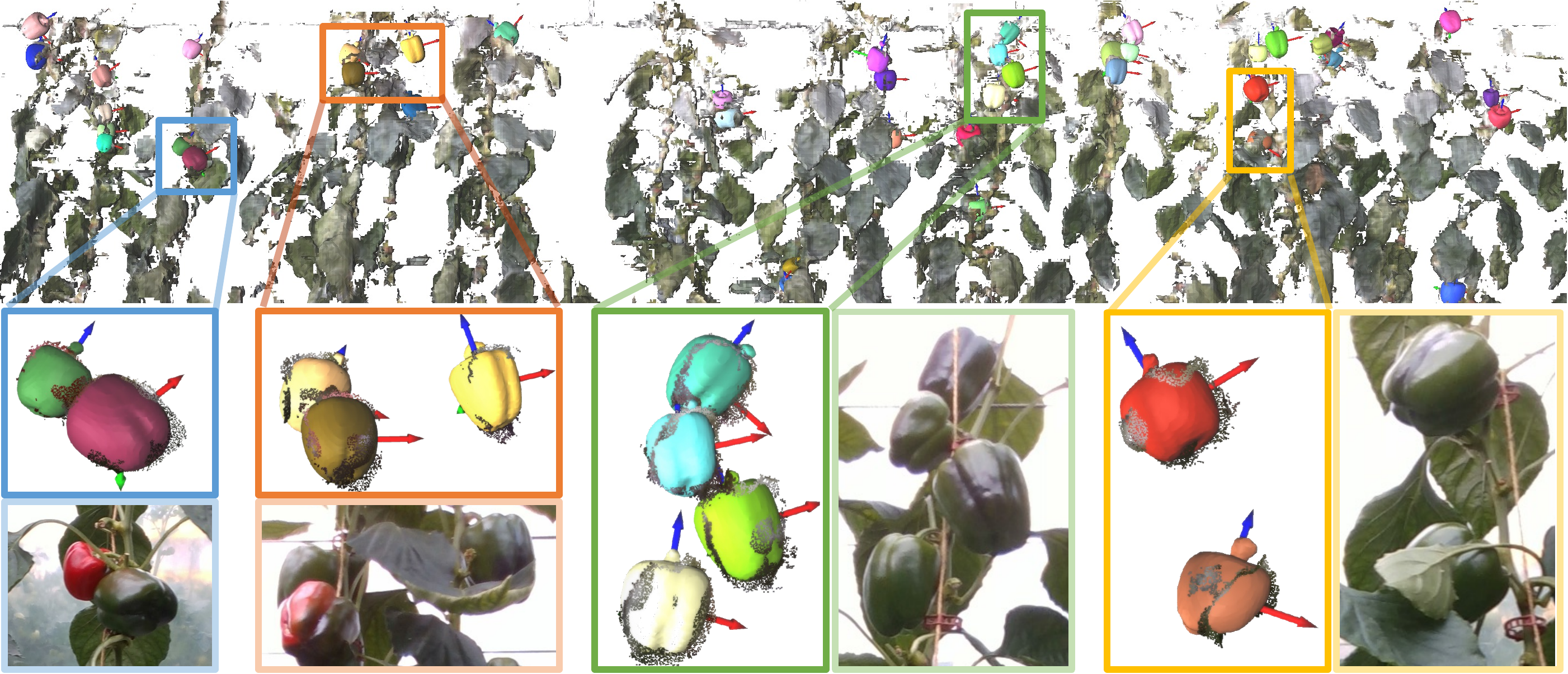}
  \caption{Qualitative results of the panoptic mapping with fruit shape completion and pose estimation in a sweet pepper greenhouse. Our method successfully achieves convincing reconstructions and poses, even with partial observations and despite heavy occlusions. We also show a failure case in the orange box on the right, where noisy partial input leads to wrong pose of the sweet pepper on the top.}
  \label{fig:row_quali}
  \vspace{-4pt}
\end{figure*}

\begin{table*}
    \centering
     \caption{Reconstruction results in the commercial greenhouse.  The $\downarrow$ and $\uparrow$ indicate that lower or higher values mean better performance.}
    \label{tab:results_wild_pepper}
    \begin{tabular}{cccccccc} 
      \toprule
      \multirow{2}{*}{\textbf{Approach}} & \textbf{$\boldsymbol{D}_\text{C}$} [mm] & \textbf{f-score} [\%] & \textbf{precision} [\%] & \textbf{recall} [\%] & \textbf{$\boldsymbol{E}_\text{rot}$} [$\deg$] & \textbf{$\boldsymbol{E}_\text{tran}$} [mm] & \textbf{inference time} [s] \\
       & $\downarrow$ avg & $\uparrow$ avg & $\uparrow$ avg & $\uparrow$ avg  &  $\downarrow$ avg &  $\downarrow$ avg &  $\downarrow$ avg\\
      \midrule
      CPD~\cite{myronenko2010pami}           & 25.38 & 3.09 & 8.10 & 1.92 &  26.79 & 27.74 & 0.57 \\    
      PF-SGD~\cite{marks2022icra}            & 9.28 & 35.03 & 37.32 & 33.21 & 29.61 & 19.73  & 30.21  \\  
      \midrule
      DeepSDF~\cite{park2019cvpr}            & 9.33 & 35.24 & 32.38 & 38.77 & \xmark & \xmark & 16.01 \\  
      CoRe~\cite{magistri2022ral-iros}       & 6.90 & 41.47 & 43.17 & 41.64 & \xmark & \xmark & \bf{0.004} \\                    
      \midrule
      Ours           &   \bf{5.29}  & \bf{58.56} & \bf{61.28} & \bf{56.26} & \bf{11.48} & \bf{11.20} & 0.62 \\  
      \bottomrule
    
    \end{tabular}
   \end{table*}

%
The main focus of this work is a pipeline for multi-resolution mapping in orchard environments including fruits shape completion and pose estimation.
%
We present our experiments to show the capabilities of our method. The results of our experiments also support the claims that our approach is able to jointly
(i) predict the 3D shape of fruits even under substantial occlusions in real commercial greenhouse environments and
(ii) estimate the pose of each fruit in the 3D map.
(iii) Additionally, our multi-resolution map representation yield substantial improvement in the shape completion task over fixed-resolution maps.

\subsection{Experimental Setup}

To showcase the capability of our proposed pipeline, we consider two typical and challenging crops, sweet pepper and strawberry, both of which have irregular shapes.

\textbf{Dataset:} For training and testing our shape completion network in a controlled environment we use the strawberries and sweet peppers dataset also used in our previous work~\cite{magistri2022ral-iros}. We, additionally, collected a sweet pepper dataset with the robot shown in \figref{fig:mot} in a greenhouse near Bonn, Germany, that we use for testing and evaluating our proposed solution in real conditions. This dataset contains \rgbd frames collected with the robot using an Intel RealSense d435i camera with a framerate of \SI{15}{\hertz}. We then harvested about 50 sweet peppers present in the greenhouse and scanned them with a high-precision handheld LiDAR system as in Schunk~\etalcite{schunck2021plosone}. In this way, after manually aligning the fruit point cloud obtained with the two different sensors, we obtain the ground truth shape of each sweet pepper and also ground truth poses with respect to the fruit canonical pose, \ie, with the peduncle pointing upwards. See \figref{fig:dataset} for a visual impression of our greenhouse dataset.

\textbf{Metrics:} 
To evaluate our shape completion solution, we use the Chamfer distance $D_\text{C}$, \ie, the average symmetric squared distance of each point to its nearest neighbor in the other point cloud.
We, additionally, use the f-score, precision, and recall at a fixed threshold ($\rho$\,=\,5\,mm in our experiments) as proposed by Knapitsch \etalcite{Knapitsch2017tog}.
To evaluate the pose estimated by our approach we report separate metrics for translations and rotations.
We report the average translation error $E_\text{tran}$, \ie, the Euclidean distance between the predicted and the ground truth center, for each fruit. We define the rotation error $E_\text{rot}$ as the the intersection angle between the \textit{z}-axis of predicted and ground truth pose. This metric ignores rotations around the fruit main axis as our target fruits are almost symmetric around it. 
Additionally, we report the average inference time needed to obtain the complete 3D shape with the estimated pose. In our experiments, we used an NVIDIA Quadro RTX A4000 GPU.

\textbf{Parameter settings:} 
We tune the hyperparameters of our system for better performance. We set the weight for each loss term in \eqref{equ:all_loss} as $w_s=1.0$, $w_d=0.05$, $w_m=0.0002$, $w_r=0.0005$ to ensure that the Hessian matrices of every loss term have the same order of magnitude. We set the damping factor $\lambda=0.1$ in \eqref{equ:lm}. To balance the efficiency and the performance, we sample $\left|\mathcal{P}_\mathcal{S}\right|=2,000$ points from each fruit submap. For each fruit instance on the image, we sample $G_f=300$ and $G_b=300$ pixels from the foreground and the background, respectively. For each ray, we sample $N=30$ points on it. We set the occlusion and surface noise thresholds to $d_o=3$\,cm and $\sigma=1$\,mm.


\begin{table*}
  \centering
   \caption{Reconstruction results in \emph{controlled environment}. The $\downarrow$ and $\uparrow$ indicate that lower or higher values mean better performance. }
  \label{tab:results_lab}
   \resizebox{\linewidth}{!}{
  \begin{tabular}{cccccc|ccccc|cc} 
    \toprule
    \multirow{3}{*}{\textbf{Approach}} & \multicolumn{5}{c|}{\normalsize{\textbf{Sweet Pepper}}}  & \multicolumn{5}{c|}{{\normalsize\textbf{Strawberry}}} & \multirow{3}{*}{\textbf{learning}} & \multirow{3}{*}{\textbf{pose}}\\
    & \textbf{$\boldsymbol{D}_\text{C}$} [mm] & \textbf{f-score} [\%] & \textbf{precision} [\%] & \textbf{recall} [\%] & \textbf{time} [s] & \textbf{$\boldsymbol{D}_\text{C}$} [mm] & \textbf{f-score} [\%] & \textbf{precision} [\%] & \textbf{recall} [\%] & \textbf{time} [s] \\
    & $\downarrow$ avg & $\uparrow$ avg & $\uparrow$ avg & $\uparrow$ avg &  $\downarrow$ avg & $\downarrow$ avg & $\uparrow$ avg & $\uparrow$ avg & $\uparrow$ avg &  $\downarrow$ avg \\
    \midrule
    CPD~\cite{myronenko2010pami}           & 12.36  & 39.84 & 76.68 & 27.07 & 15.62 & 5.13 & 57.93 & 94.09 & 42.34  & 0.57 & \xmark & \cmark \\    
    PF-SGD~\cite{marks2022icra}            & 3.97 & 68.95  & 71.20 & 66.94  & 17.48 & 2.71  & 86.08  & 88.82  & 83.90  & 8.10 & \xmark & \cmark \\  
    \midrule
    DeepSDF~\cite{park2019cvpr}            & 29.78  & 37.12  & 32.96  & 46.06 & 44.13 & 3.61  & 74.01  & 83.76  & 68.32 & 36.84 & \cmark & \xmark \\  
    CoRe~\cite{magistri2022ral-iros}       &  7.83 & 52.85  & 47.38  & 60.00  & \bf{0.004} & 2.67  & 86.01 & 87.97  & 84.85  & \bf{0.004} & \cmark & \xmark \\    
    \midrule
    Ours                            & \bf{3.16} & \bf{80.86}  & \bf{82.14}  & \bf{79.72}  & 0.60  & \bf{2.42} & \bf{92.81} & \bf{94.38} & \bf{91.53} & 0.53 & \cmark & \cmark \\  
    \bottomrule
  
  \end{tabular}
  }
 \end{table*} 

\begin{table*}
    \centering
     \caption{Results of different maps in the greenhouse. The $\downarrow$ and $\uparrow$ indicate that lower or higher values mean better performance.}
    \label{tab:results_wild_pepper_maps}
     \resizebox{\linewidth}{!}{
    \begin{tabular}{cccccccc|c} 
      \toprule
      \multirow{2}{*}{\textbf{Approach}} & \textbf{$\boldsymbol{D}_\text{C}$} [mm] & \textbf{f-score} [\%] & \textbf{precision} [\%] & \textbf{recall} [\%] & \textbf{$\boldsymbol{E}_\text{rot}$} [$\deg$] & \textbf{$\boldsymbol{E}_\text{tran}$} [mm] & \textbf{inference time} [s] & \multirow{2}{*}{\textbf{Online Map}}\\
       & $\downarrow$ avg & $\uparrow$ avg & $\uparrow$ avg & $\uparrow$ avg  &  $\downarrow$ avg &  $\downarrow$ avg &  $\downarrow$ avg\\
      \midrule
      Bundle Adjustment           & \bf{4.70} & \bf{67.71} & \bf{69.07} & \bf{66.52} & \bf{10.23} & \bf{8.71} & 0.66 & \xmark \\     
      \midrule
      RGB-D Single Frame                     & 9.62 & 40.56 & 41.98 & 39.52 & 25.91 & 18.21 & \bf{0.16} & \cmark\\  
      Fixed-Resolution Map                   & 7.29 & 44.22 & 46.29 & 42.62 & 17.48 & 16.02 & 0.75 & \cmark\\  
      Multi-Resolution Map (Ours)           &  5.29  & 58.56 & 61.28 & 56.26 & 11.48 & 11.20 & 0.62 & \cmark\\  
      \bottomrule
    
    \end{tabular}
    }
    \vspace{-1em}
   \end{table*}



\subsection{Shape Completion and Pose Estimation}
The first experiment evaluates the performance of our approach and supports the claims that our approach can predict the 3D shape of fruits
under substantial occlusions and can estimate the pose of each fruit in the 3D map. The qualitative results are shown in~\figref{fig:row_quali}.
We report in~\tabref{tab:results_wild_pepper} the metrics regarding shape completion and fruit pose estimation in real greenhouses. Our approach yields better performances in all metrics except for the inference time where we can still estimate poses and shapes of two fruit per second. The most competitive baselines for shape completion~\cite{magistri2022ral-iros, park2019cvpr} provide Chamfer distance performances of 1.6\,mm  and 4.0\,mm worse than our approach and can not estimate fruit poses, thus limiting their applicability in this scenario. On the other hand, the best baseline that can estimate both shape and pose provide a Chamfer distance of about 4\,mm, a rotation error of 10$^{\circ}$ and a translation error of about 8\,mm worse than ours.

For a deeper comparison with the baselines and for testing our approach on a different fruit species, we evaluate the shape completion accuracy in a controlled environment in the laboratory where we estimate shapes of sweet peppers and strawberries, see~\tabref{tab:results_lab}. Our approach yields better reconstruction performances on both species with an f-score of 80.86\% on the sweet pepper and of 92.81\% on the strawberry, while the closest baseline~\cite{marks2022icra} reaches 68.95\% and 86.08\% respectively. In terms of execution time, our approach can estimate the 3D shape of roughly two fruit per second, note that the baseline~\cite{magistri2022ral-iros} with lower inference time produce worse 3D shapes.




\subsection{Influence of Map Representation on Completion Results}
The second experiment evaluates the effects of the map representation on estimating shapes and poses of fruits. This experiment illustrates that our multi-resolution map representation yield substantial improvement in the shape completion task over fixed-resolution maps. In~\tabref{tab:results_wild_pepper_maps}, we compare the shape completion results obtained with using our multi-resolution mapping strategy ($1\,\mathrm{cm}$ / $3\,\mathrm{mm}$ voxel size) against the results of fixed-resolution maps ($1\,\mathrm{cm}$ voxel size) and single \rgbd frames. Note that our multi-resolution mapping system can run in real-time. We additionally report the shape completion results obtained with a map obtained offline using bundle-adjustment software.  For the Chamfer distance, our map yields a 28\% improvement over fixed-resolution map and a 45\% improvement over single \rgbd frame. Considering the poses, using our map yields an improvement of 6$^{\circ}$ and 5\,mm  with respect to fixed-resolution map and an improvement of 14$^{\circ}$ and 7\,mm for over single \rgbd frame. Notably, by using our online mapping strategy, we obtain competitive results with respect to using an offline map built by photogrammetric bundle adjustment. Our completion results are only 0.6\,mm less accurate than the one obtained by using the offline method. Similarly, our predicted pose is around 1$^{\circ}$ and 3\,mm less accurate.

\subsection{Ablation Study}
We ran several ablation studies to evaluate the impact of each loss term in~\eqref{equ:all_loss}. From the metrics in~\tabref{tab:ablation_loss}, it is clear that our loss design with all four terms yields the best performances in both shape completion and pose estimation. As expected, the regularization term $\mathcal{L}_\mathrm{r}$ has the lowest impact in the final results. Instead, the biggest impact on the performances is given by the mask term $\mathcal{L}_\mathrm{m}$ defined in~\eqref{equ:mask}. Without this term, the f-score drops to 35.89\% instead of 58.56\%, the rotation error increases from 11.48$^{\circ}$ to 19.16$^{\circ}$ and the translation error from 11.20\,mm to 23.34\,mm.

\begin{table}
  \caption{Ablation studies on the loss terms} 
  \label{tab:ablation_loss}
  \centering
  \resizebox{0.95\linewidth}{!}{
  \begin{tabular}{ccccc} 
    \toprule
    \multirow{2}{*}{\textbf{Loss}} & \textbf{$\boldsymbol{D}_\text{C}$} [mm] & \textbf{f-score} [\%] & \textbf{$\boldsymbol{E}_\text{rot}$} [$\deg$] & \textbf{$\boldsymbol{E}_\text{tran}$} [mm]  \\
     & $\downarrow$ avg  & $\uparrow$ avg & $\downarrow$ avg  & $\downarrow$ avg  \\
    \midrule
    No $\mathcal{L}_\mathrm{s}$                           & 8.17 & 38.71 & 15.80 & 17.99 \\
    No $\mathcal{L}_\mathrm{d}$                         & 8.93 & 36.12 & 17.11 & 19.78  \\
    No $\mathcal{L}_\mathrm{m}$                           & 10.09 & 35.89 & 19.16 & 23.34 \\
    No $\mathcal{L}_\mathrm{r}$                          & 7.64 & 40.24 & 15.42 & 12.34 \\
    All Loss (Ours)                                      & \bf{5.29} & \bf{58.56} & \bf{11.48} & \bf{11.20} \\
    \bottomrule
  \end{tabular}}
  \vspace{-1em}
\end{table}

\section{Conclusion}
\label{sec:conclusion}

We presented a novel approach for panoptic mapping in real agricultural greenhouses using a mobile robot equipped with \rgbd cameras. We build a multi-resolution map to represent fruits with higher resolution and jointly estimate complete 3D shapes of fruits and their pose in the 3D map. Our method exploits high-precision 3D scanning to learn a general fruit shape prior that we use at inference time together with an occlusion-aware differentiable rendering pipeline. This allows us to successfully complete partial fruit observations and estimate the 7 DoF pose of each fruit in the map. We solve the joint shape and pose optimization efficiently with analytical Jacobians, allowing for its application online. We evaluated our approach on different datasets and provided comparisons to other existing techniques and supported all our claims made. The experiments suggest that the proposed approach yields higher shape completion and pose estimation accuracy than existing baselines. It provides precise and complete fruits models, allowing farmers to assess the status of the orchard. Furthermore, our map providing fruits shape and pose explicitly can serve as a basis for planning autonomous fruit harvesting missions.



\bibliographystyle{plain_abbrv}

\bibliography{glorified,new}

\end{document}